\documentclass[10pt, letterpaper, journal, twoside]{IEEEtran}

\IEEEoverridecommandlockouts    
   
\usepackage{algorithm} 

\usepackage{graphicx}
\usepackage{tikz}

\usepackage{graphics} 
\usepackage{cite}
\usepackage[T1]{fontenc}
\usepackage{amsmath, bm} 
\usepackage{amssymb}  

\usepackage[colorlinks=false]{hyperref}

\usepackage{float}
\usepackage{cancel}

\usepackage{numprint}
\usepackage{comment}

\usepackage{color, colortbl}
\definecolor{LightCyan}{rgb}{0.88,1,1}
\usepackage{multirow}
\usepackage{siunitx}
\usepackage{array}
\DeclareSIUnit\cell{cell}
\DeclareSIUnit\cells{cells}
\DeclareSIUnit\trees{trees}
\usepackage{amsmath,mathtools}
\usepackage[nodisplayskipstretch]{setspace}
\usepackage{titlesec} 
\renewcommand{\thesection}{\Roman{section}}
\renewcommand{\thesubsection}{\thesection-\Alph{subsection}}
\renewcommand{\thesubsubsection}{\thesubsection\arabic{subsubsection}}
\titleformat{\subsubsection}[runin]{\itshape}{\arabic{subsubsection})}{0.5em}{}

\titlespacing*{\subsubsection}{\parindent}{0pt}{*1}
\titlespacing*{\subsection}{0pt}{\baselineskip}{20pt}
\titlespacing*{\section}{0pt}{\baselineskip}{20pt}

\titlespacing*{\section}{0pt}{*1}{*1}
\titlespacing{\subsection}{0pt}{*1}{*1}

\usepackage{enumitem} 

\usepackage{wrapfig}

\usepackage{tikz}
\usetikzlibrary{shapes,arrows}
\usepackage{verbatim}
\usetikzlibrary{positioning}
\usepackage[normalem]{ulem} 
\usepackage{tabstackengine}
\usepackage{algorithm}
\usepackage[end]{algpseudocode} 
\usepackage{etoolbox}
\usepackage[font=small, skip=5pt]{caption}
\usepackage[skip=3pt]{subcaption} 

\usepackage{pgfplots}
\usepackage{grffile}
\pgfplotsset{compat=newest}
\usetikzlibrary{plotmarks}
\usetikzlibrary{arrows.meta}
\usepgfplotslibrary{patchplots}
\usepackage{amsmath}

\usepackage{bbding} 

\usepackage{colortbl}

\usepackage{hhline}
\usepackage{makecell}

\makeatletter
\newcommand*\titleheader[1]{\gdef\@titleheader{#1}}
\AtBeginDocument{%
  \let\st@red@title\@title
  \def\@title{%
    \bgroup\normalfont\large\centering\@titleheader\par\egroup
    \vskip1.5em\st@red@title}
}
\makeatother

\title{\LARGE \bf Autonomous Mapless Navigation on Uneven Terrains}

\author{Hassan Jardali, Mahmoud Ali, and Lantao Liu
\thanks{Authors are with the Luddy School of Informatics, Computing, and Engineering, Indiana University, Bloomington, IN 47408 USA (e-mail: {\tt\small \{hjardali, alimaa, lantao\}@iu.edu}). \\
}
}%
\definecolor{applegreen}{rgb}{0.8, 1, 0.0}
\definecolor{LightCyan}{rgb}{0.88,1,1}
\definecolor{atomictangerine}{rgb}{1.0, 0.6, 0.4}
\definecolor{amber}{rgb}{1.0, 0.75, 0.0}
\definecolor{aqua}{rgb}{0.0, 1.0, 1.0}
\definecolor{almond}{rgb}{0.94, 0.87, 0.8}
\definecolor{aquamarine}{rgb}{0.5, 1.0, 0.83}
\definecolor{babyblue}{rgb}{0.54, 0.81, 0.94}
\definecolor{babyblueeyes}{rgb}{0.63, 0.79, 0.95}
\definecolor{asparagus}{rgb}{0.53, 0.66, 0.42}
\definecolor{auburn}{rgb}{0.43, 0.21, 0.1}
\definecolor{brilliantlavender}{rgb}{0.96, 0.73, 1.0}
\definecolor{bittersweet}{rgb}{1.0, 0.44, 0.37}
\definecolor{blue-violet}{rgb}{0.54, 0.17, 0.89}
\definecolor{capri}{rgb}{0.0, 0.75, 1.0}
\definecolor{celadon}{rgb}{0.67, 0.88, 0.69}
\definecolor{darkcyan}{rgb}{0.0, 0.55, 0.55}
\definecolor{deepskyblue}{rgb}{0.0, 0.75, 1.0}
\definecolor{dogwoodrose}{rgb}{0.84, 0.09, 0.41}

\titleheader{ \textcolor{gray}{This paper has been accepted for publication at 2024 IEEE International Conference on Robotics and Automation} \textcolor{blue}{\href{https://2024.ieee-icra.org/}{(ICRA 2024)}} }

\begin{document}

\maketitle

\global\csname @topnum\endcsname 0
\global\csname @botnum\endcsname 0


\thispagestyle{empty}
\pagestyle{empty}


\begin{abstract}
We propose a new method for autonomous navigation in uneven terrains by utilizing a sparse Gaussian Process (SGP) based local perception model. The SGP local perception model is trained on local ranging  observation (pointcloud) to learn the terrain elevation profile and extract the feasible navigation subgoals around the robot. 
Subsequently, a cost function, which prioritizes the safety of the robot in terms of keeping the robot's roll and pitch angles bounded within a specified range, is used to select a safety-aware subgoal that leads the robot to its final destination.
The algorithm is designed to run in real-time and is intensively evaluated in simulation and real-world experiments. The results compellingly demonstrate that our proposed algorithm consistently navigates uneven terrains with high efficiency and surpasses the performance of other planners. The code can be found here: \href{https://rb.gy/3ov2r8}{https://rb.gy/3ov2r8}.


\end{abstract}


\vspace{3pt}
\section{Introduction}\label{Introduction}




Autonomous navigation in outdoor unstructured environments poses a significant challenge due to unpredictable, rugged terrains. Advanced navigation algorithms, utilizing sensors like LiDARs and cameras, have been developed to tackle these issues, enabling robots to navigate complex and challenging terrains more precisely and in real-time.
In order to accomplish autonomous navigation, various modules such as sensing, perception, localization, planning, and control are essential. Our main contribution is a GP-based local perception model capable of extracting feasible, traversable navigation points from the robot's immediate environment.
To date, the approaches used to navigate challenging terrains can be categorized as map-based methods and mapless methods, where for the map-based methods, a global map is built based on which the path planing techniques are utilized to drive the robot; and for mapless methods the robot uses only its current sensor reading to reach a pre-specified goal without referring to any global map.
%
Mapless navigation offers several advantages such as less computational power as it obviates the necessity for continuous map updates. It also facilitates rapid response to dynamic obstacles and alterations in the environment.

We have recently developed a novel method to process LiDAR pointclouds by converting them into a  "truncated" {\em Occupancy Surface} where the pointclouds received within a pre-specified radius are projected onto this surface~\cite{ali2023light}. 
The occupancy surface represents a more compressed form of the environment which enables an efficient communication of the robot's local observation.
Additionally, this occupancy surface was used in navigation and exploration as demonstrated in \cite{ali2023exp,ali2023gp,mohamed2023gp}.
This occupancy surface serves as training data for a  GP occupancy model, which predicts a continuous form of occupancy around the robot.
In this paper, we leverage the uncertainty inherent in the GP occupancy model to estimate the traversability of the uneven terrain and identify local navigation points.
Subsequently, a cost function is applied to evaluate these navigation points, selecting the most promising point for guiding the robot towards its final destination as the subsequent navigation subgoal. This selection process takes into account constraints on the robot's orientation, ensuring that they remain within its safe bounds.
Our approach has been verified through rigorous simulations, comparative benchmarks and real-world demonstrations. The results show that our proposed method enables the robot to perform efficient mapless navigation in uneven terrains.

\vspace{3pt}
\section{Related Work}\label{Related Work}



Autonomous navigation frameworks can be generally categorized into two paradigms: sense-plan-act and end-to-end. The first involves mapping, traversability analysis, planning, and control, whereas the second employs learning-based approaches like deep learning and reinforcement learning \cite{zhou2022autonomous}. In this study, we focus on methodologies within the sense-plan-act paradigm.
Terrain traversability analysis assesses the difficulty of a vehicle might face when navigating a particular terrain section. It aids robots in choosing the most optimal route between multiple possible paths \cite{9738557}. Grid-based, 2.5D elevation \cite{8392399} and 3D OctoMap \cite{hornung2013octomap} mapping methods are common approaches to solve the map representation problem. In \cite{kim2018slam}, a multi-layer 2D map is obtained from a 3D OctoMap then employed to navigate staircases and slopes, however, its effectiveness is constrained in navigating intricate environments \cite{9981038}. \cite{8392399} introduced a real-time elevation mapping technique that mitigates localization drift in legged robots. This approach creates a probabilistic terrain estimate using a grid-based elevation map, where each cell provides an estimated height along with confidence boundaries. The studies detailed in \cite{7759199,9738557} leverage elevation maps to craft strategies for guiding legged robots safely and efficiently across irregular, unstructured terrains. Additionally, in \cite{9738557}, terrain material is classified using learning methods based on RGB data, which is subsequently integrated in the traversability analysis as a navigability criterion. These frameworks extract geometric properties from the terrain's elevation information within each grid cell, producing a traversability score that is logged in a traversability cost map. This score, derived from the terrain's geometric features, is computed by considering factors such as roughness, slope, sparsity, step height, and flatness \cite{7759199,9738557,5354535,9981038,10160330}. Path planning algorithms such as A* search \cite{4082128, 10160330}, Dijkstra's algorithm \cite{ammar2016relaxed}, and the rapidly-exploring random tree (RRT) algorithm \cite{lavalle2001randomized}, are then used to generate the path after building the map. A motion command is finally sent to the robot's controller after the path is generated. Optimization methods are sometimes included in the controller such as in \cite{10160330, 9981038}. In \cite{10160330}, the authors introduce a distinctive field construction algorithm in 3D space, named the Valid Ground Filter (VGF), that processes the original map taking into account the robots navigation capabilities. In \cite{9981038}, 3D grid map is used to build the map, along PF-RRT*, an RRT variant based on the plane fitting method to include the traversability information while generating the path. The framework in \cite{9981038} depends also on a SLAM module, which makes it prone to cumulative errors and difficult to scale on large maps. 



Navigation systems that operate without reliance on maps,
frequently utilize reactive behavior methodologies such as the dynamic window approach (DWA) ~\cite{580977}, the vector field histogram (VFH) \cite{borenstein1991vector}, and Artificial Potential Fields (APF) \cite{khatib1986real} approaches. Although these methods were originally devised for 2D navigation, advancements in robotics and research have seen their extension to 3D scenarios \cite{peng2014obstacle}. 

Unlike the methods discussed earlier, our approach to autonomous navigation in uneven terrains relies on a novel lightweight perception modeling technique that utilizes the uncertainty of the GP. By using LiDAR readings as training data, the GP model is used to create a representation of the robot's surroundings. The GP uncertainty is then utilized to obtain the terrain elevation profile and identify local navigation subgoals. These subgoals are evaluated based on a cost function that drives the robot towards the final goal while ensuring its safety by imposing penalties on steep paths.

\section{Preliminaries}\label{GP_prelim}
Our local perception model is based on Gaussian Process (GP). In this section, we provide preliminary background of GP. GP is a non-parametric model defined by a mean function $m(\textbf{z})$ and a covariance function $k(\textbf{z}, \textbf{z}^{\prime})$, which is known as GP  kernel, where $\textbf{z} \in \mathbb{R}^{d}$ represents the input~\cite{GPforML}, 
\begin{equation}
    f(\mathbf{z}) \sim \mathcal{G P}\left(m(\mathbf{z}), k\left(\mathbf{z}, \mathbf{z}^{\prime}\right)\right).
    \label{eq_full_gp}
\end{equation} 

In GP regression, we train a GP model on dataset $D = \left\{\left(\mathbf{z}_{i}, y_{i}\right)\right\}_{i=1}^{n}$, where $y_i=f(\textbf{z}_i)+\epsilon_i$, with $f(\textbf{z}_i)$ is the underlying function and  $\epsilon_i \sim \mathcal{N}(0,\sigma^2)$ is Gaussian noise. Then, the output $y^*$ of a new input $\textbf{z}^*$ is calculated using the GP posterior mean and co-variance functions, $m_{\mathbf{y}}(\mathbf{z})$ and $k_{\mathbf{y}}(\textbf{z},\textbf{z}^{\prime})$, respectively, as follow 
\begin{equation}
    \begin{aligned}
    p(y^* | \textbf{y}) &= \mathcal{N}(y^* | m_{\textbf{y}}(\textbf{z}^*), k_\textbf{y}(\textbf{z}^*,\textbf{z}^*) + \sigma^2), \\
        m_{\mathbf{y}}(\mathbf{z}) &=K_{\mathbf{z} n}\left(\sigma^{2} I+K_{n n}\right)^{-1} \mathbf{y}, \\
    k_{\mathbf{y}}\left(\mathbf{z}, \mathbf{z}^{\prime}\right) &=k\left(\mathbf{z}, \mathbf{z}^{\prime}\right)-K_{\mathbf{z} n}\left(\sigma^{2} I+K_{n n}\right)^{-1} K_{n \mathbf{z}^{\prime}},
     \end{aligned}
    \label{eq_predictive_eq_full_gp}
\end{equation} 
where $K_{\mathbf{z}n}\in \mathbb{R}^n$ is $n$-dimensional row vector of kernel function values between $\mathbf{z}$ and the training inputs, with $K_{n\mathbf{z}} = K_{\mathbf{z}n}^T$, while $K_{nn} \in \mathbb{R}^{n \times n}$ refers to the $n \times n$ co-variance matrix of the training inputs. 
Maximizing the logarithm of the marginal likelihood, described in eq.~\ref{eq:eq_mlml_full_gp}, is a common approach to optimizing hyper-parameters, including kernel parameters $\Theta$ and noise variance $\sigma^2$, which in turn leads to more accurate GP predictions.
\begin{equation}
    \log p(\mathbf{y})=\log \left[\mathcal{N}\left(\mathbf{y} \mid \mathbf{0}, \sigma^{2} I+K_{n n}\right)\right].
    \label{eq:eq_mlml_full_gp}
\end{equation} 
The computational cost of the Gaussian Process (GP), which is $\mathcal{O}(n^3)$, can be prohibitive in some applications. To address this issue, Sparse Gaussian Process (SGP) has emerged as an effective alternative. By using a smaller set of training points, known as the "inducing points" $Z_{m}$, SGP achieves lower computational complexity of $\mathcal{O}(n m^2)$ ~\cite{lawrence2003fast, snelson2006sparse, titsias2009variational}. 
In this work, we utilize the Variational SGP (VSGP) approach introduced in~\cite{titsias2009variational2}, which estimates the true posterior of GP $p(f|\mathbf{y})$ with an approximated variational posterior distribution $q(f,f_{m})$ by introducing an unconstrained variational distribution $\phi(f_{m})$ as follow
\begin{equation}
    \begin{aligned}
    p(f,f_{m}|\mathbf{y}) = p(f|f_{m}) p(f_{m}|y), \\
    q(f,f_{m}) = p(f|f_{m})\phi(f_{m}),
     \end{aligned}
    \label{eq_approx_equal_true_posterior_SGP}
\end{equation}
 where $f_{m}$ are the values of $f(\mathbf{z})$ at $Z_{m}$ and $p(f|f_{m})$ is the conditional GP prior.
Using VSGP, the inducing inputs $Z_{m}$ and hyperparameters ($\Theta, \sigma^2$) are estimated by minimizing the Kullback-Leibler (KL) divergence between the true and approximated posteriors, $\mathbb{K} \mathbb{L}[q(f, f_{m})||p(f|\mathbf{y})]$.

\begin{figure*}[ht!] 
\subfloat[\label{fig_init_sgp_oc_b}]{%
  \includegraphics[width=0.19\textwidth,height=1.15in]{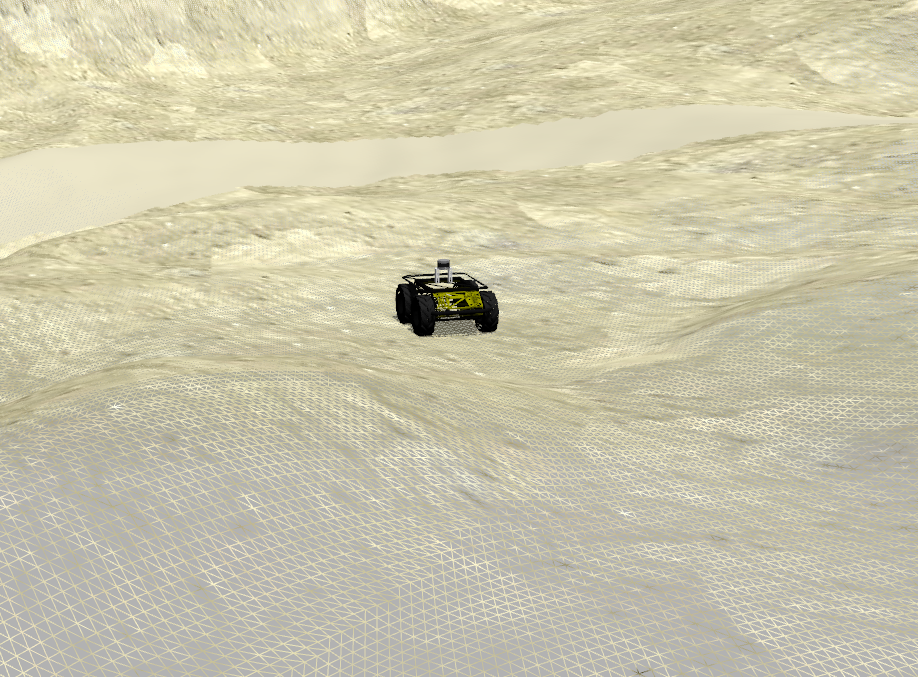} } \hfill
\subfloat[\label{fig_sgp_oc_f}]{%
  \includegraphics[width=0.19\textwidth,height=1.15in]{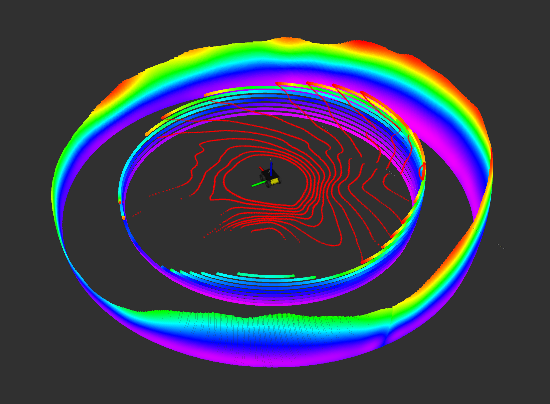}}  \hfill
\subfloat[\label{fig_sgp_oc_g}]{%
  \includegraphics[width=0.19\textwidth,height=1.15in]
  {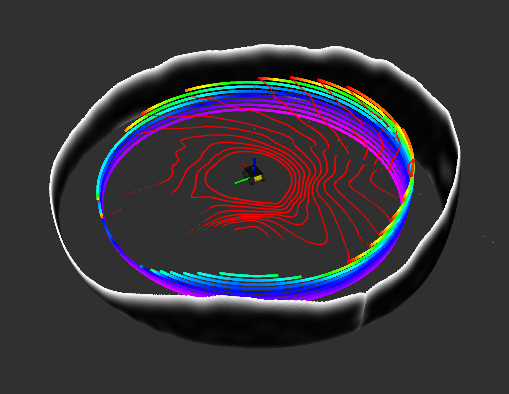} }
  \includegraphics[width=0.01\textwidth,height=1.15in]{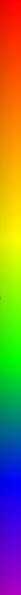} \hfill
\subfloat[\label{fig_sgp_oc_h}]{%
  \includegraphics[width=0.19\textwidth,height=1.15in]{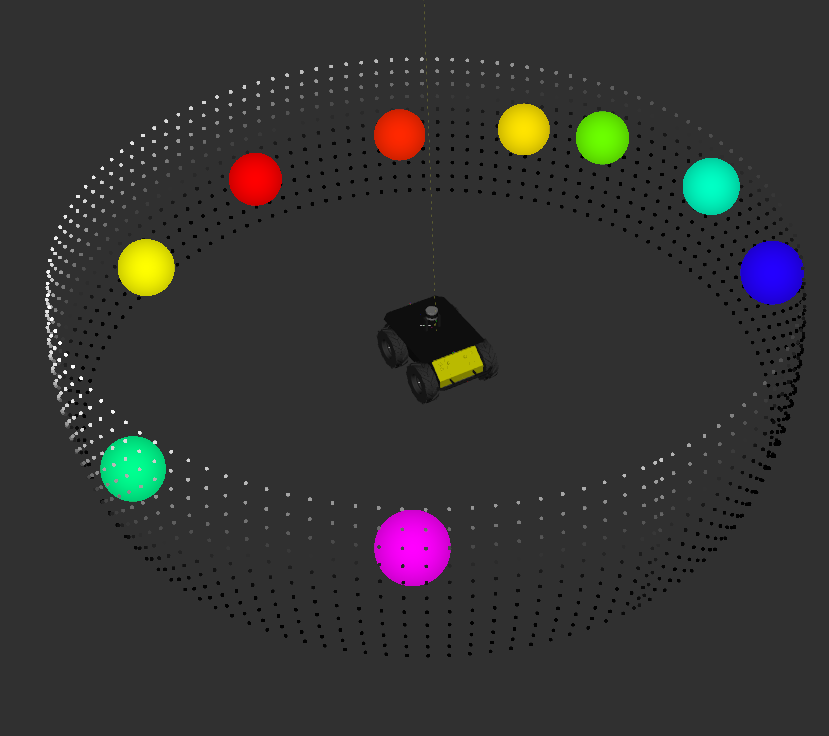}}\hfill 
\subfloat[\label{fig_sgp_oc_h2}]{%
  \includegraphics[width=0.19\textwidth,height=1.15in]{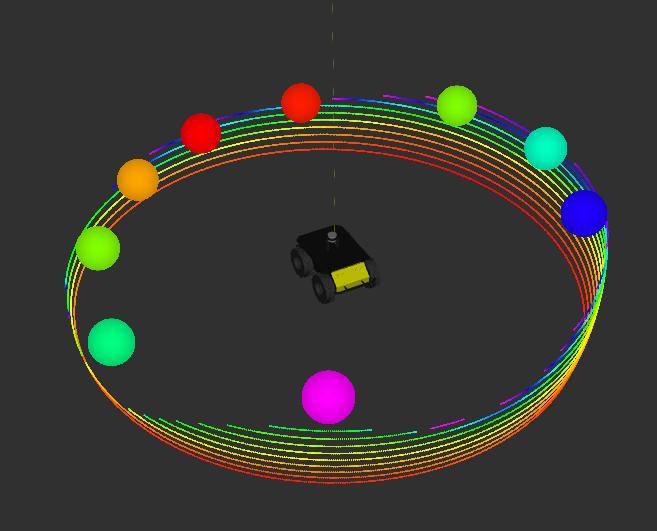}}\hfill 
  \caption{\small   
  (a) Husky robot in a simulated hilly terrain environment; (b) Original (inner) vs predicted (outer) occupancy surfaces, where warmer colors indicate less occupancy; (c) Original occupancy (inner) vs. variance (outer) surface; (d) The full variance surface with subgoals shown as colored circles; (e) The original occupancy surface with the extracted subgoals. \vspace{-10pt} }
  \label{fig_sgp_oc}
\end{figure*}

\section{Problem Definition}\label{Problem_Definition}
Consider a  differential wheeled mobile robot with a 
motion model given as
\begin{equation*}
\begin{bmatrix}
\dot{x} \\
\dot{y} \\
\dot{\theta}
\end{bmatrix}
=
\begin{bmatrix}
\cos\theta & 0\\
\sin\theta & 0\\
0 & 1
\end{bmatrix}
\begin{bmatrix}
v \\
\omega \\
\end{bmatrix} ,
\end{equation*}
with $\mathbf{x}=(x, y, \theta)$ is the robot pose, where $\theta$ is the robot's rotation angle around the z-axis (yaw); and $\mathbf{u}=(v, \omega)$ is the control input, where $v$ is the linear velocity and $\omega$ is the angular velocity. 

Let $\mathbf{x}_s=(x_s, y_s, \theta_s)$ and $\mathbf{x}_f=(x_f, y_f, \theta_f)$ denote the robot's initial and final poses, respectively. The robot has to move from $\textbf{x}_s$ to $\textbf{x}_f$ through $\mathbb{R}^3$, where $\mathbb{R}^3$ represents the 3D space of the global environment where the robot moves, with $\mathcal{S}_{n} \subset \mathbb{R}^3$ and $\mathcal{S}_{nn} \subset \mathbb{R}^3$ representing the global navigable and non-navigable spaces, respectively.

The navigation task can be described as finding the optimal path that guides the robot to move from a starting pose, $\mathbf{x}_s$, to a final goal $\mathbf{x}_f$, safely and efficiently while avoiding obstacles and minimizing costs (e.g. traveled distance, robot's pitch and roll angles, etc). 
In the context of terrain navigation, safety means preventing the robot from flipping or skidding by keeping the roll $\phi$ and pitch $\psi$ angles, robot's rotation angles around the $X$-axis and $Y$-axis, respectively, in a safe bounded range such as $ \phi_{min} \leq \phi < \phi_{max}$ and $ \psi_{min} \leq \psi < \psi_{max}$. 

Considering only the local perception, we simplify this task by only considering the local space around the robot $\mathcal{S}_l^3 \subset SE(3)$, with $\mathcal{S}_{l_n} \subset \mathcal{S}_l^3$ and $\mathcal{S}_{l{_{nn}}} \subset \mathcal{S}_l^3$ representing the local navigable and non-navigable spaces, respectively. Subsequently, the local navigable space $\mathcal{S}_{l_n}$ is used to extract a set of feasible subgoals $\mathcal{G}$
around the robot. 
Then, an optimal subgoal $g^*$ is selected based on a cost function $\textbf{J}$, which minimizes the distance and direction of the robot with respect to the target's pose $\textbf{x}_f$. Finally, a control input $\textbf{u}$ is generated to drive the robot towards a subgoal $g^*$.

\section{Methodology}\label{methodology}
In this section, we 
initiate our discussion with the Gaussian Process based perception model, then we cover the process of subgoals generation, and concluding with the formulation of the cost function used to select the most optimal subgoal. 
\begin{figure*}[t] 
    \centering
      \includegraphics[width=0.99\textwidth,height=0.63in]{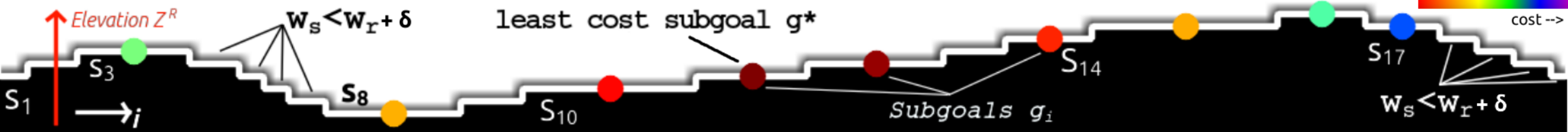} 
  \caption{\small Unfolded variance surface transformed into a 2D image, where dark areas indicate the location of the observed points, and the white areas above the dark area represent the free space. The horizontal segments $\textbf{S}\{s_i\}_{i=1}^Q$ reflect the discretized terrain's elevation profile of the scene shown in Fig. \ref{fig_sgp_oc} as seen from the robot's sensor. 
  $Z^{R}_{s_i}$ is the height of a segment $s_i$ in the robot's frame $\mathcal{R}$. A set of subgoals $\mathcal{G}=\{g_i\}_{i=1}^K$ is placed at the centers of the corresponding segments whose width satisfies $w_{s_i}\!<\!w_r + \delta$. The subgoals' colors indicate the cost assigned to each subgoal $g_i$ by the cost function $\textbf{J}$, where warmer colored subgoals have less cost.\vspace{-10pt}}\label{var_img} 
\end{figure*}
\subsection{Perception Representation using Gaussian Process}\label{VSGP_oc_mdl}
This paper extends our previous work~\cite{ali2023light} which represents the observed pointcloud as {\em occupancy surface}, then models it using the SGP. The occupancy surface is a 2D circular surface around the sensor center with a predefined radius $r_{oc}$. Higher $r_{oc}$ values indicate the compression of a greater amount of terrain information along each LiDAR beam. 
The points of the observed pointcloud are transformed into the spherical coordinates $(\alpha_i, \beta_i, r_i)$ and projected on the occupancy surface, where $(\alpha_i, \beta_i, r_i)$ correspond to the azimuth, elevation, and radius values, respectively.
Any point $\mathbf{z}_i$ on the surface is described by the azimuth and elevation angles $\mathbf{z}_i= (\alpha_i, \beta_i)$, and given an {\em occupancy value} $f(\mathbf{z}_i)$, where $f(\mathbf{z}_i)=r_{oc}-r_i$. 
Thus, the occupancy distribution over the surface, $f(\mathbf{z})$, is modeled as an SGP occupancy model, 
\begin{equation}
    \begin{aligned}
    f(\mathbf{z}) &\sim \mathcal{SGP}\left(m(\mathbf{z}), k\left(\mathbf{z}, \mathbf{z}^{\prime}\right)\right),  \\
    k\left(\mathbf{z}, \mathbf{z}^{\prime}\right) &=\sigma^{2}\left(1+\frac{\left(\mathbf{z}-\mathbf{z}^{\prime}\right)^{2}}{2 \gamma \ell^{2}}\right)^{\!\!-\gamma}\!\!,
    \end{aligned}
    \label{eq_mean_kernel_SGP}    
\end{equation}
where $k\left(\mathbf{z}, \mathbf{z}^{\prime}\right)$ is a Rational Quadratic (RQ) kernel with a length-scale $l$, a signal variance $\sigma^{2}$, and a length-scale weighting factor $\gamma$. The RQ kernel is used because of its capacity to model functions that vary across diverse length-scales~\cite{GPforML}.
The zero-mean function $m(\mathbf{z})=0$ sets the prior occupancy of the surface to zero.

The prediction of the SGP occupancy model consists of a mean value indicating the predicted occupancy and a variance value revealing the uncertainty of the predicted occupancy at each point on the occupancy surface.
Hence, the outcome of the SGP occupancy model consists of two surfaces, the predicted (reconstructed) occupancy surface and the variance (uncertainty) surface, which indicates the uncertainty of the predicted occupancy, see Fig.~\ref{fig_sgp_oc}. 
The LiDAR does not cast any pointclouds if the laser beams do not hit any obstacle; hence, the absence of pointclouds causes high variance in the corresponding areas in the variance surface. Therefore, free space areas can be conveniently defined as high variance areas where the variance is higher than a certain threshold $V_{th}$, a tunable parameter that is related to the mean of the variance distribution over the variance surface.
\vspace*{-5pt}

\subsection{Subgoals Placement}
\vspace*{-5pt}
The variance surface offers valuable insights into the morphology of the terrain's around the robot. In flat environments~\cite{ali2023gp}, the variance surface reveals the distribution of obstacles and identifies free spaces. In uneven terrains, the variance surface displays the varying elevations of the terrain, as depicted in Fig.~\ref{var_img}. We opted to utilize the variance surface as it more smoothly and distinctly delineates occupied and free spaces, facilitating the thresholding process and thus the mapping of local terrain elevations.
Specifically, Fig.~\ref{var_img} is an unfolded version of the circular variance surface (the out-most surface) shown in Fig.~\ref{fig_sgp_oc_g}, where dark regions (lower part) represent occupied space (terrain), while white regions (upper part) indicate the free space. 
The border between the occupied and free spaces is represented by an uneven curve.
This curve is discretized into a set of segments, $\textbf{S}=\{s_i\}_{i=1}^{Q}$, where each segment $s_i$ reflects an elevation level of the terrain around the robot. 
Each segment $s_i$ is defined by two azimuth angles, $\alpha_{sb_i}$ and $\alpha_{se_i}$, which indicate the beginning and the ending points of the segment, respectively, where the segment width $w_{s_i}$ is calculated as $w_{s_i}=r_{oc} (\alpha_{se_i}-\alpha_{sb_i})$. 
Additionally, each segment has an elevation angle $\beta_{s_i}$ coupled with a height value, $z^{R}_{s_i}$, which indicates the terrain elevation in the robot's frame $\mathcal{R}$, see Fig.~\ref{var_img}. 
The segments $\textbf{S}$ are used to place a set of potential local navigation subgoals, $\mathcal{G}\!=\!\{g_i\}_{i=1}^{K}$ around the robot, where the segment height $z^{R}_{s_i}$ is used to calculate the steepness cost $C_{stp}$ of the corresponding subgoal $g_i$.

A segment $s_i$ is considered for subgoal positioning \textit{if and only if}: 
i) the segment width $w_{s_i}$ is greater than the robot width augmented by a safety margin, i.e., $w_{s_i} > w_r + \delta$, where $\delta$ is a constant representing the safety threshold; and 
ii) the segment height $z^{R}_{s_i}$ falls within a safe slope range, which is determined by the robot's climbing capabilities influenced by its dynamics.
For each segment \( s_i \) that satisfies the previously mentioned conditions, we initially place a subgoal \( g_i \) at the \textit{center} of the segment. This placement is achieved using the spherical coordinates in \( \mathcal{R} \), denoted as \( g_i = (\alpha_{g_i}, \beta_{g_i}, r_{oc}) \), where \( \alpha_{g_i} = (\alpha_{se_i} - \alpha_{sb_i})/2 \) and \( \beta_{g_i} = \beta_{s_i} \). If a segment has a width \( w_s \) exceeding \( n \cdot (w_r + \delta) \), where \( n \geq 3 \), we distribute \( n-1 \) subgoals at equidistant points along the segment to facilitate a smoother path toward the global goal \( x_f \).
The cartesian coordinates of a subgoal $g_i$, in the global frame  $\mathcal{W}$, is retrieved as $(x^{w}_{g_i}, y^{w}_{g_i}, z^{w}_{g_i}) = \prescript{W}{}{\textbf{T}}_{R} (x^{R}_{g_i}, y^{R}_{g_i}, z^{R}_{g_i})$, where $\prescript{W}{}{\textbf{T}}_{R}$ is the transformation between $\mathcal{R}$ and $\mathcal{W}$ (robot localization), and $(x^{R}_{g_i}, y^{R}_{g_i}, z^{R}_{g_i})$ are the cartesian coordinates of the subgoal $g_i$ in $\mathcal{R}$. 


\begin{figure}[t] 
\centering
\includegraphics[height=0.35\linewidth, width=0.9\linewidth]{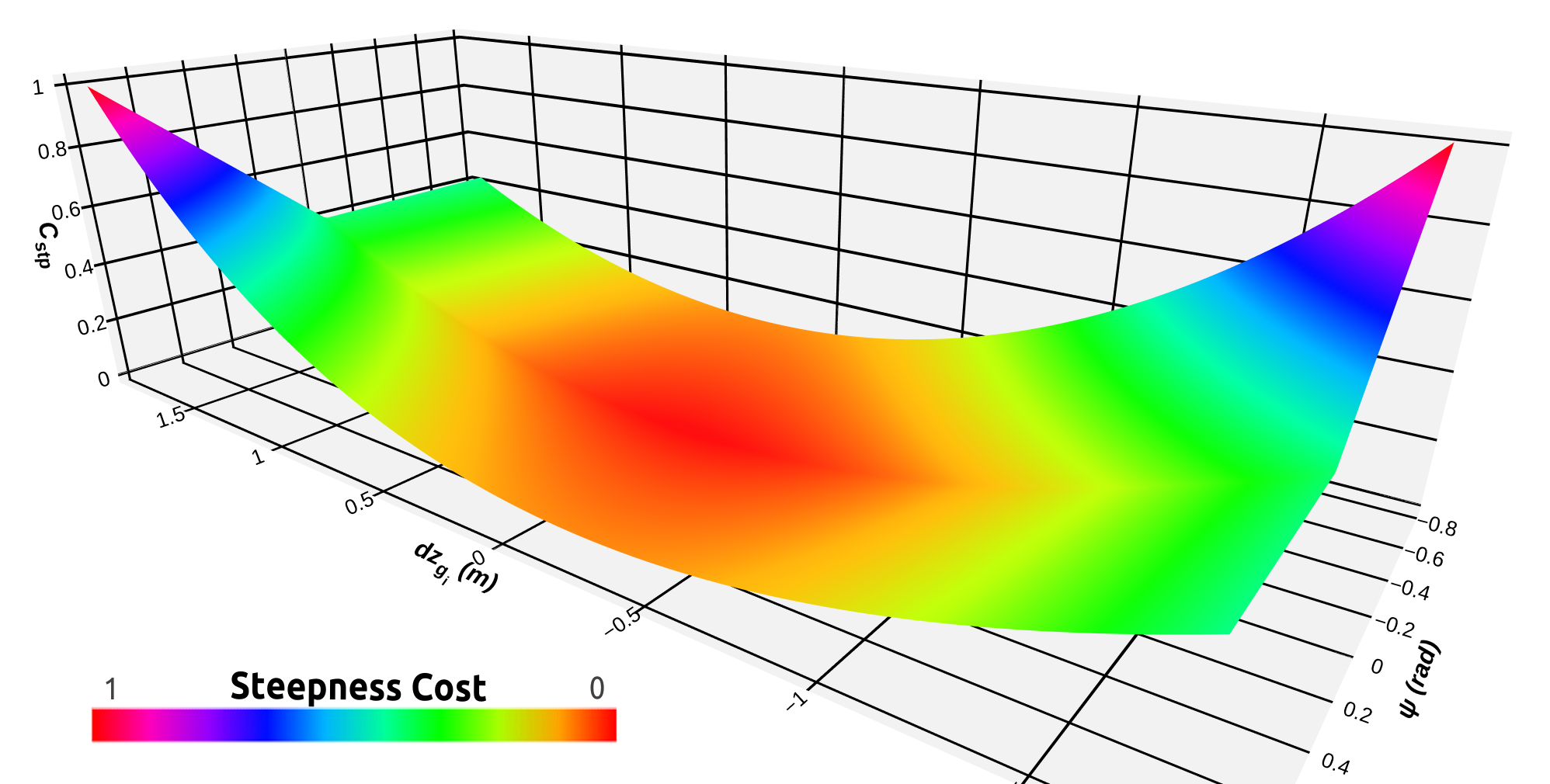}
\caption{\small This graph illustrates the behavior of the steepness cost function \(\mathbf{C}_{\text{stp}}\) with respect to the pitch angle \(\psi\) and the relative elevation of subgoals (\(dz_{g_i}\)). For negative \(\psi\) (indicating the robot is inclined upwards), the cost is higher when \(dz_{g_i}\) is negative (subgoals are higher than the robot) compared to when \(dz_{g_i}\) is positive (subgoals are lower than the robot). Conversely, for positive \(\psi\) (indicating the robot is inclined downwards), the function assigns a lower cost to negative \(dz_{g_i}\) values and a higher cost to positive \(dz_{g_i}\) values. This structure ensures the containment of roll and pitch angles within safe limits.}
\label{fig_c_stp}
\vspace*{-20pt}
\end{figure}
\subsection{Selection of Optimal Feasible Subgoal}
After the placement of the subgoals $\mathcal{G}\!=\!\{g_i\}_{i=1}^{k}$, a cost function $\textbf{J}$ is used to evaluate each subgoal $g_i$. In \cite{ali2023gp}, the distance and the direction between the subgoal $g_i$ and final goal $\textbf{x}_f$ were used as factors to drive the robot in flat terrains. 
For the uneven terrain case, we first check the feasibility of the subgoal based on the steepness of the terrain using the steepness cost function $\textbf{C}_{stp}$ in Eq.~\eqref{eq_j}, which depends on the elevation of the goal with respect to the robot, $dz_{g_i} = z^w_{g_i} - z^w_r$, and the robot's pitch angle $\psi$, where $z^w_{g_i}$ and $z^w_r$ are the Z-coordinates of the subgoal $g_i$ and the robot in $\mathcal{W}$, respectively. 
An exponential \textit{pitch-cost}, $\exp(\textit{sign}(\psi)dz_{g_i})|\psi|$, is used to penalize the subgoals that will increase the absolute value of the robot's pitch angle, and a quadratic \textit{elevation-change} cost, $dz_{g_i}^2$, to penalize the subgoals that lead to greater elevation changes $\dot{z_r}$, see Fig.~\ref{fig_c_stp}. We add to the steepness cost a distance cost \( \textbf{D}_{fg_i} \)~\cite{yan2020mapless} and a direction cost \( \alpha_{g_i} \)~\cite{ye2009sub}. The \( \textbf{D}_{fg_i} \) represents the distance between each subgoal \( g_i \) and the final goal \( \textbf{x}_f \), and \( \alpha_{g_i} \) represents the azimuth angle of the subgoal \( g_i \) around the Z-axis. This angle is defined as the difference in heading between the robot's current orientation and the direction towards $g_i$, measured in the robot's frame. The three costs are normalized and combined in one cost function $\textbf{J}$ as follows
\begin{equation}
\begin{aligned}
\textbf{J}\left(g_{i}\right) &= k_{dir} \alpha_{g_i} + k_{dst} \textbf{D}_{fg_i} +  k_{stp}  \textbf{C}_{stp} ,  \\
 \textbf{D}_{fg_i}\left(g_{i}\right) &= ( (x^w_f - x^w_{g_i})^2 + (y^w_f - y^w_{g_i})^2 )^{\frac{1}{2}},\\
\textbf{C}_{stp}\left(g_{i}\right) &= (dz_{g_i})^2 +\exp( \textit{sign}(\psi)dz_{g_i})|(\psi)|, \\
g^{*} &=\operatorname{arg} \min _{g_{i} \in \mathcal{G}}\left(\textbf{J}\left(g_{i}\right)\right), \\
\end{aligned}
    \label{eq_j}    
\end{equation}
where coefficients $k_{dst}$, $k_{stp}$, and $k_{dir}$ can be adjusted through empirical tuning to account for the terrain's flatness and steepness while also considering the dynamics of the robot. 
The subgoal with the minimum cost $g^*$ is selected as the next navigation subgoal. Finally, a control input $\textbf{u}$ is generated to drive the robot to $g^*$ using a simple PID controller with a distance error 
equals to the distance between the selected subgoal $g^*$ and the robot, and a yaw-orientation error of $\alpha_{g^*}$.


\section{Simulation-Based Evaluation}\label{Simulation-Based Evaluation}
Our method is coded in Python using \textit{gpflow}~\cite{matthews2017gpflow} and integrated with the Robot Operating System (ROS) framework. It runs in \textit{real-time} with a frequency of 10 Hz on a \texttt{core-i7} PC with Nvidia RTX-2060 GPU.

\subsection{Simulation Setup}\label{Simulation Setup}
We tested our approach using a simulated \textit{Husky} robot equipped with a 16-beam Velodyne LiDAR in Gazebo simulator. 
The LiDAR is used to construct the occupancy surface, which spans an azimuth range of $-180^o \leq \alpha < 180^o$ and an elevation angle range of $-15^o\leq\beta<15^o$. 
The occupancy surface radius $r_{oc}$ is set to 7 meters, while the variance threshold $V_{th}$ is set to 0.7.
For the simulation experiments, the weighting factors of the cost function are set as the direction factor $k_{dir}=0.2$, the distance factor $k_{dst}=0.3$, and the steepness factor $k_{stp}=0.5$.

\subsection{Performance Metrics and Simulation Scenarios}\label{Simulation Setup}
The performance of the proposed method is evaluated through extensive simulations and compared to a very recent uneven-terrain navigation method~\cite{10160330}, referred here as Valid Ground Filter (VGF). This planner addresses the navigation in uneven terrain using a map-based strategy. The outcomes highlighted in the article demonstrated its superior performance over other methods, and the authors have made it open source.
Two sets of performance metrics are considered for uneven terrain navigation: \textit{safety-related} metrics and \textit{motion-related} metrics.  
 The \textit{safety-related} metrics encompass the robot's roll $\phi$ and pitch $\psi$ angles, the change in elevation $\dot{z}_r$ during the robot's movement from $\textbf{x}_s$ to $\textbf{x}_f$, and the vibration $\mathbf{v}_{b}$, as defined in \cite{liang2022adaptiveon} as the cumulative rate of change for the roll and pitch angles: $\mathbf{v}_{b}= \lvert\omega_{r} \rvert+ \lvert\omega_{p} \rvert$, with $\omega_{r}=\dot{\phi}$ and $\omega_{p}=\dot{\psi}$. A lower vibration value is indicative of a more stable trajectory.
 The \textit{motion-related} metrics include the distance $D$ taken to achieve the final goal and the average trajectory curvature change ${C}_{chg}$ as in \cite{mujahed2017admissible}, representing the path's oscillations,
$$
{C}_{chg}=\frac{1}{{T}} \int_0^{T}\left|k^{\prime}(t)\right| d t, \quad k(t)=\left|\frac{\omega(t)}{v(t)}\right|.
$$
A superior performance is denoted by reduced values for both safety and motion-related metrics outlined above.
To ensure the robot's safety, we adhere to the safety boundaries defined by the robot's manufacturer, \textit{Clearpath Robotics} in our case, which stipulates that the vehicle tends to slide or skid when its roll exceeds \(0.524 \, \text{rad}\) or its pitch exceeds \(0.785 \, \text{rad}\). 
Hence, we define a trial as successful if the robot reaches the final goal and its roll and pitch angles remain within the aforementioned constraints; otherwise, the trial is considered a failure.
Furthermore, achieving the final destination, $\textbf{x}_r\!=\!\textbf{x}_f$ is a requisite for a successful simulation. 
Three distinct simulated environments were used to evaluate the performance of our approach:
\begin{enumerate}
    \item \textbf{The Challenging Hilly Terrain (CHT) Environment}: A versatile environment offering three distinct navigable paths, each with its own set of challenges. 
    \item \textbf{The Martian Landscape (ML) Environment}: Resembling the terrain of the Mars planet.
    \item \textbf{The Valley Trail Ascent (VTA) Environment}: This environment portrays a valley leading to a trail that ascends to a higher point on a small mountain. 
\end{enumerate}


    
    


\subsection{Results}\label{Simulation Results}
\vspace*{-15pt}
\begin{figure}[htb]
    \begin{subfigure}{.48\columnwidth}
        \centering
        \includegraphics[width=1\linewidth, height=0.95\linewidth]{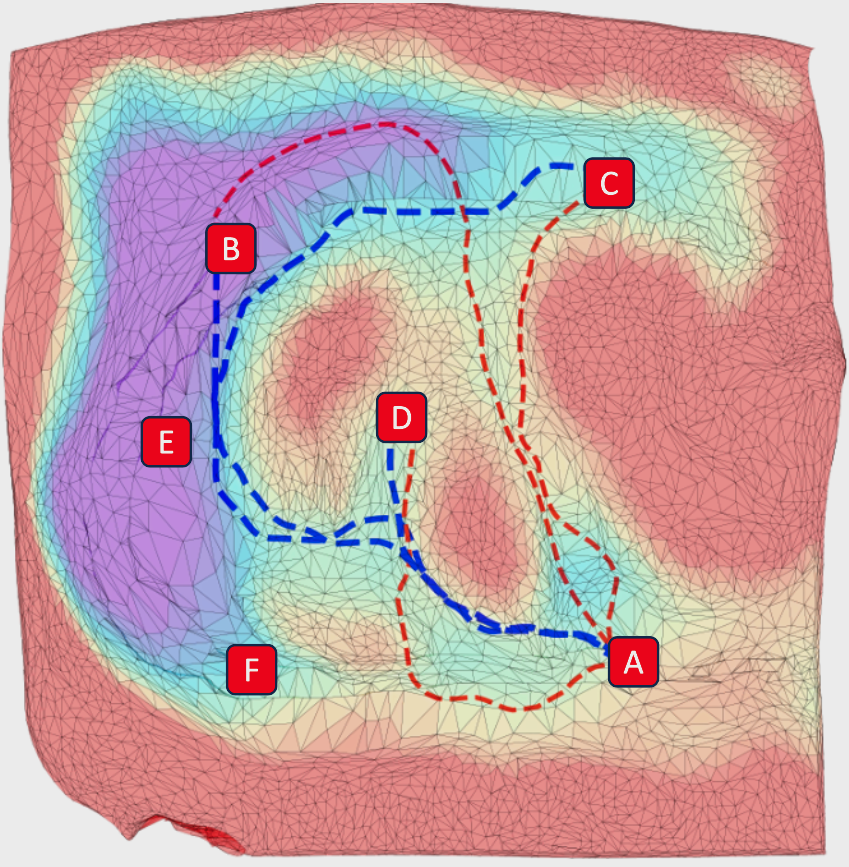}
        \caption{\small }
        \label{fig:scenario1}
    \end{subfigure}%
    \hspace{0.01\columnwidth} 
    \begin{subfigure}{.48\columnwidth}
        \centering
        \includegraphics[width=1\linewidth, height=0.95\linewidth]{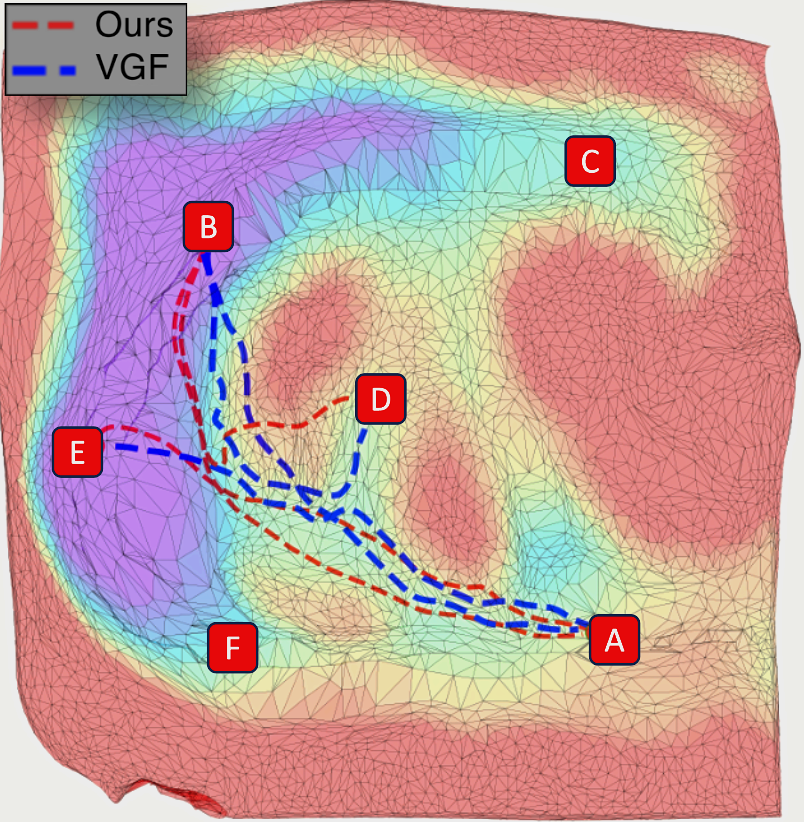}
        \caption{\small }
        \label{fig:scenario2}
    \end{subfigure}

    \caption{\small Path comparison in the CHT environment between our proposed approach and the baseline. The color variations represent the terrain elevation, with red shades denoting higher areas.}

    \label{fig:path_comparison}
    \vspace{-15pt}
\end{figure}

\begin{figure}[htb]
    \begin{subfigure}{.48\columnwidth}
        \centering
        \includegraphics[width=1\linewidth, height=0.7\linewidth]{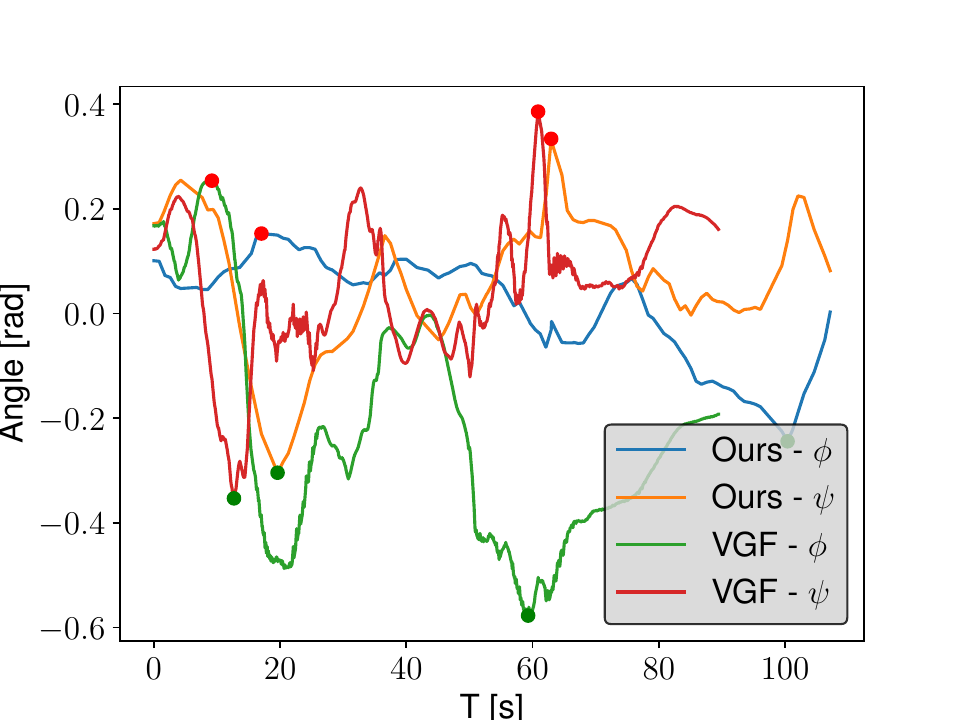}
        \caption{\small $(\phi)$ and $(\psi)$ angles variation}
        \label{fig:rollpitcha2b}
    \end{subfigure}%
    \hspace{0.01\columnwidth} 
    \begin{subfigure}{.48\columnwidth}
        \centering
        \includegraphics[width=1\linewidth, height=0.7\linewidth]{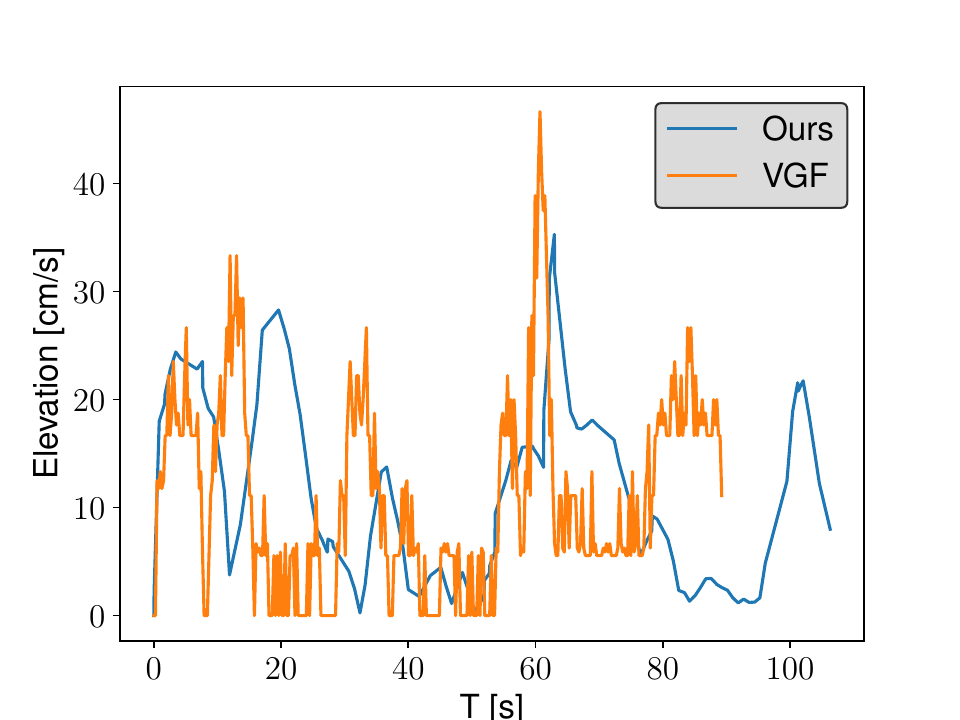}
        \caption{\small 
        Elevation change \(\dot{z}_r\) profile
        }
        \label{fig:elevationa2b}
    \end{subfigure}

    \caption{\small Illustrations of roll and pitch angle variations and elevation changes during the AB path trials, as depicted in \ref{fig:scenario1}.}
    \label{fig:metric_comparison}
\end{figure}


\begin{table*}[h!]
\centering
{
\small
\setlength{\tabcolsep}{9pt}
\begin{tabular}{|c|c|c|c|c|c|c|c|}
\hline
\textbf{Experiment} & \textbf{Method} & \textbf{\(D\) (m)} & \textbf{\(|\phi|_{\text{max}}\) (rad)} & \textbf{\(|\psi|_{\text{max}}\) (rad)} & \textbf{\(\mathbf{v}_{b_{avg}}\) (m/s)} & \textbf{\( \dot{z}_{r_{avg}} \) (cm/s)} & \textbf{\(C_{chg}\) (rad/m)} \\
\hline
\multirow{2}{*}{AD} & Ours & 63.456 & \textbf{0.444} & \textbf{0.282} & \textbf{0.0617} & \textbf{5.18} & 0.096 \\
& VGF & \textbf{46.566} & 0.489 & 0.355 & 0.1175 & 9.04 & \textbf{0.0234} \\
\hline
\multirow{2}{*}{AE} & Ours & 74.816 & 0.492 & \textbf{0.354} & \textbf{0.0702} & \textbf{10.39} & 0.0693 \\
& VGF & \textbf{73.929} & \textbf{0.488} & 0.475 & 0.1104 & 12.49 & \textbf{0.0236} \\
\hline
\multirow{2}{*}{BA} & Ours & 91.860 & \textbf{0.506} & \textbf{0.242} & \textbf{0.0391} & \textbf{6.49} & 0.0732 \\
& VGF & \textbf{85.369} & 0.508 & 0.526 & 0.1067 & 13.29 & \textbf{0.0241} \\
\hline
\multirow{2}{*}{AC} & Ours & \textbf{71.901} & \textbf{0.505} & \textbf{0.278} & \textbf{0.057} & \textbf{8.18} & 0.0797 \\
& VGF & 140.487 & 0.649 & 0.362 & 0.1065 & 9.43 & \textbf{0.0246} \\
\hline
\multirow{2}{*}{BD} & Ours & 60.4 & \textbf{0.52} & \textbf{0.53} & \textbf{0.094} & \textbf{11.07} & 0.110 \\
& VGF & \textbf{60.3} & 0.69 & 0.721 & 0.172 & 17.23 & \textbf{0.029} \\
\hline
\end{tabular}
\caption{\small Comparison of metrics between our method and the baseline for selected experiments. Bold values indicate better performance.}
\label{tab:experiment_comparison}
}
\vspace{-10pt}
\end{table*}


The performance of our method is evaluated in the CHT environment against the VGF baseline~\cite{10160330} through hundreds of simulation trials.  Fig.~\ref{fig:path_comparison} illustrates the paths pursued by both methods, delineating the starting and ending points. Fig.~\ref{fig:metric_comparison} exhibits the angle and elevation variations along the AB path. 
In Table~\ref{tab:experiment_comparison}, where a quantitative comparison of the results derived from a selection of trials is presented, we can notice that our proposed method consistently exhibited lower elevation changes \( \dot{z}_r \), reduced vibration \(\mathbf{v}_{b}\), and smaller maximum absolute angles for both roll (\(\phi\)) and pitch (\(\psi\)), reflecting an enhanced emphasis on safety. However, in terms of curvature change \(C_{chg}\), the map-based approach yielded superior outcomes, a logical result considering the ability of map-based methods to provide optimized global paths.
There were instances where the baseline paths closely skirted steep hills, as illustrated in Fig.~\ref{fig:path_comparison}, leading to significant roll (\(\phi\)) values, like in the AC and AB paths scenario, consequently, the robot encountered multiple failures. For the AB scenario, although our proposed method opted for a lengthier trajectory, it experienced reduced maximum angles compared to the baseline as depicted in \ref{fig:rollpitcha2b}. Moreover, as seen in \ref{fig:elevationa2b}, our approach demonstrated a lower rate of elevation change. For the BD path, while the baseline attempted to bypass the hill, it encountered high angles values. In contrast, our method ascended the hill via navigable terrain, resulting in a safer traversal experience.
We should emphasize that our analysis was strictly based on the global paths produced by the baseline method, prior to the implementation of the MPC. The data we gathered was obtained from the Husky robot as it navigated the generated waypoints using a PID controller.

It is worth pointing out that a major difference between the two methods is that our method is mapless, while the baseline relies on a map-based strategy. The baseline method initially requires the map to be provided as pointcloud data, followed by analysis and planning using A*. This approach can yield more robust paths. Conversely, our algorithm operates with minimal prior information, limited to the global goal's relative position to the starting point. This characteristic renders our algorithm computationally lightweight and highly adaptable to variations in terrain. The objective of our comparison is to highlight the areas where our proposed method can outperform map-based planning approaches. 

\begin{figure}[htb]
    \begin{subfigure}{.45\columnwidth}
        \centering
        \includegraphics[width=0.95\linewidth,height=0.99in]{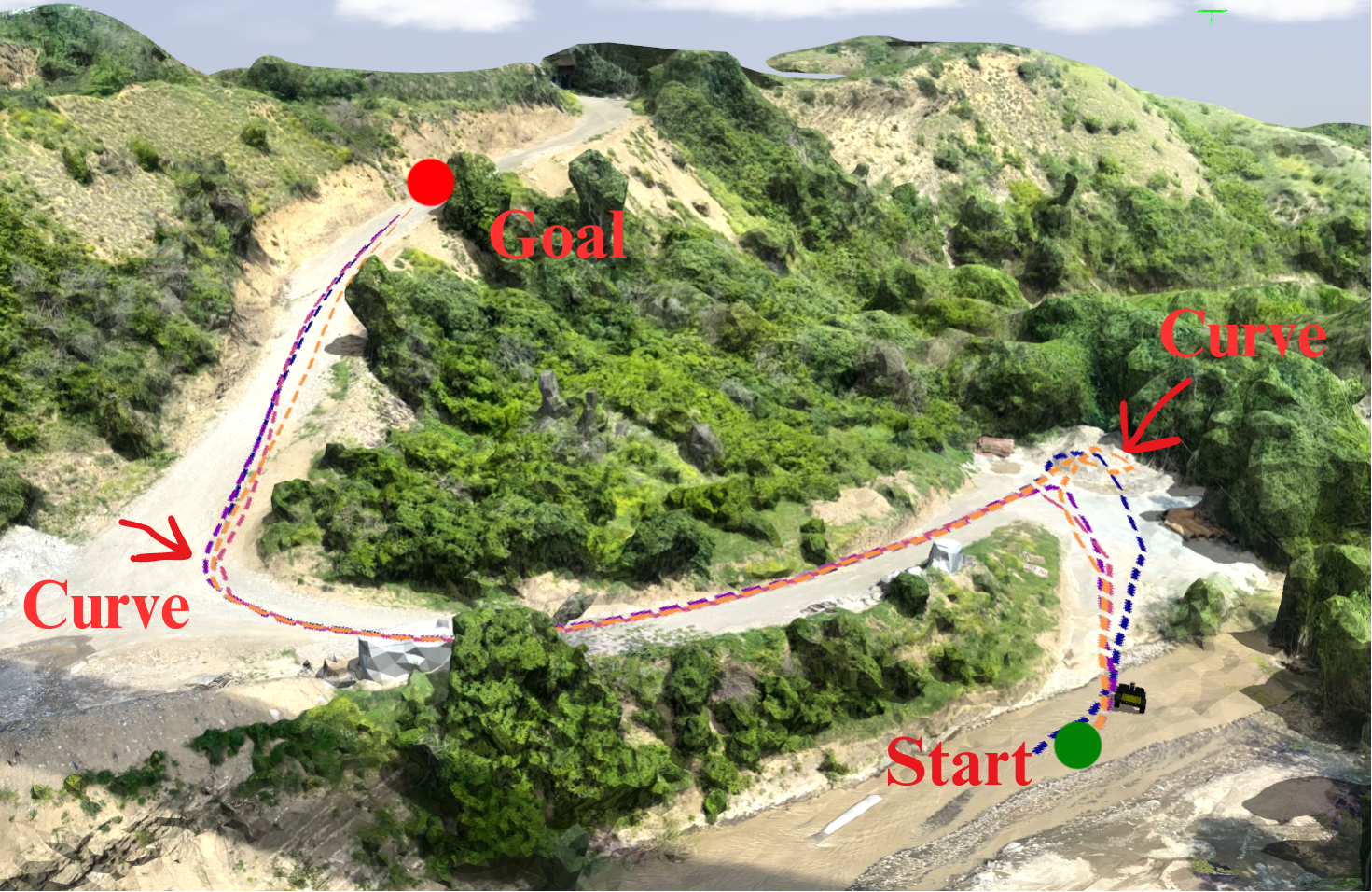}
        \caption{\small Valley Trail Ascent (VTA)}
        \label{fig:valley}
    \end{subfigure}
    \quad
    \begin{subfigure}{.45\columnwidth}
        \centering
        \includegraphics[width=0.95\linewidth,height=0.99in]{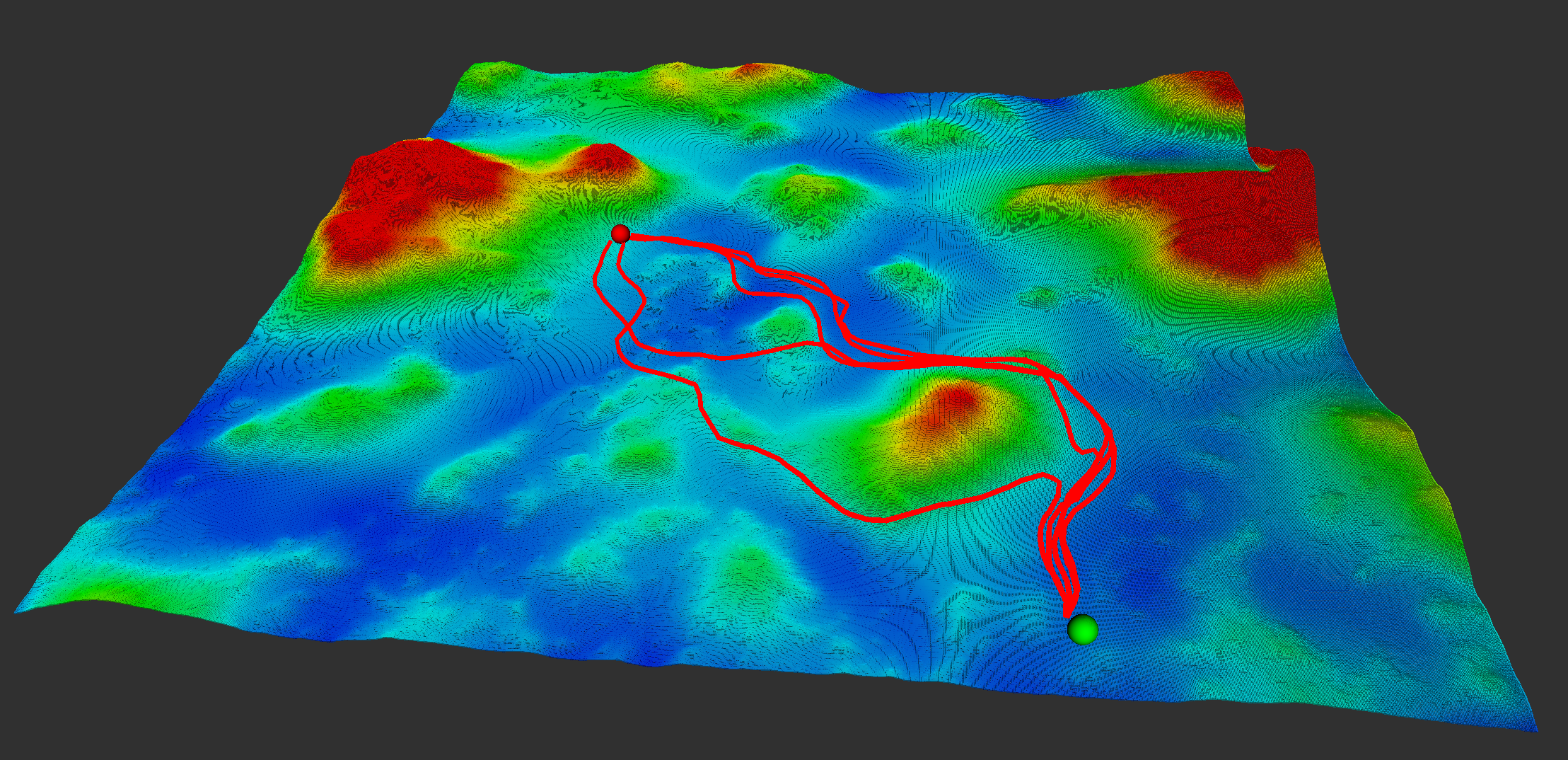}
        \caption{\small Martian Landscape (ML)}
        \label{fig:martian}
    \end{subfigure}%
    \caption{\small Simulation environments used to validate our proposed methodology. Some of the paths are shown in these figures.}
    \label{fig:simulation_environments}
    \vspace{-5pt}
\end{figure}


\begin{table}[!ht]
\begin{center}
\caption{\small \small Performance statistics in \textbf{VTA} and \textbf{ML} environments.}
\label{sim_res}
\small\addtolength{\tabcolsep}{2pt} 
\begin{tabular}{|c|c|c|}
 \hline
 \textbf{Metric} & \textbf{VTA} & \textbf{ML}  \\ 
 \hline
 \(D [m]\) & \(64.962\pm5.207\) & \(84.470\pm2.315\) \\
 \(\max|\phi| [rad]\) & \(0.246\pm0.075\) & \(0.414\pm0.087\) \\
 \(\max|\psi| [rad]\) & \(0.280\pm0.065\) & \(0.345\pm0.068\) \\
 \(\mathbf{v}_{b_{avg}} [rad/s]\) & \(0.050\pm0.007\) & \(0.071\pm0.014\) \\
 \(\dot{z}_{r_{avg}} [cm/s]\) & \(8.5\pm1.6\) & \(4.0\pm0.6\) \\
 \(C_{\text{chg}} [rad/m]\) & \(0.108\pm0.013\) & \(0.112\pm0.032\) \\
 \hline
\end{tabular}
\end{center}
\vspace{-15pt}
\end{table}

For a more comprehensive demonstration, we tested our algorithm in two distinct scenarios illustrated in Fig.~\ref{fig:simulation_environments} and in Table~\ref{sim_res}, we summarize the performance metrics from these simulation experiments. For the \textit{VTA} environment, the goal $\textbf{x}_f$ is positioned atop a hill with the starting point $\textbf{x}_s$ nestled in a valley. The navigable route is illustrated in Fig.~\ref{fig:valley}. The paths taken averaged an elevation change rate of $8.5 cm/s$, attributed to the hill-valley configuration. The key challenge in this map revolves around bypassing two pronounced curves. These curves can often lead to failures in mapless algorithms by causing the solution to oscillate between two local navigation goals, which may result in the robot colliding with an obstacle situated between the paths. 
The steepness and distance components of the cost function are key in addressing these challenges, with further metrics validating the safety of the selected paths.

Turning our attention to the \textit{ML} scenario, the average traveled distance, denoted as \(D\), is 84.47m. Notably, the maximum angles encountered were 0.41 rad and 0.35 rad.
The average of elevation change \(\dot{z}_{r_{avg}}\) was a modest 4 cm/s, signifying effective hill avoidance by the robot. This aligns with the fact that the safest path between the start and goal points in this terrain does not exhibit significant elevation variations. The curvature metric was registered at $0.112 rad/m$, slightly elevated due to the numerous hills the robot circumvented, as depicted in Fig.~\ref{fig:martian}. 
Additionally, each iteration of our algorithm exhibited a latency of approximately $0.11\,s$, encompassing the interval from receiving the point clouds to sending the command to the controller.

\vspace{-5pt}
\begin{figure}[ht!] 
\centering
\subfloat[Gentle Slope Bypass environment (GSB)\label{gentle_slope_bypass}]{%
  \includegraphics[width=0.45\textwidth,height=0.85in]{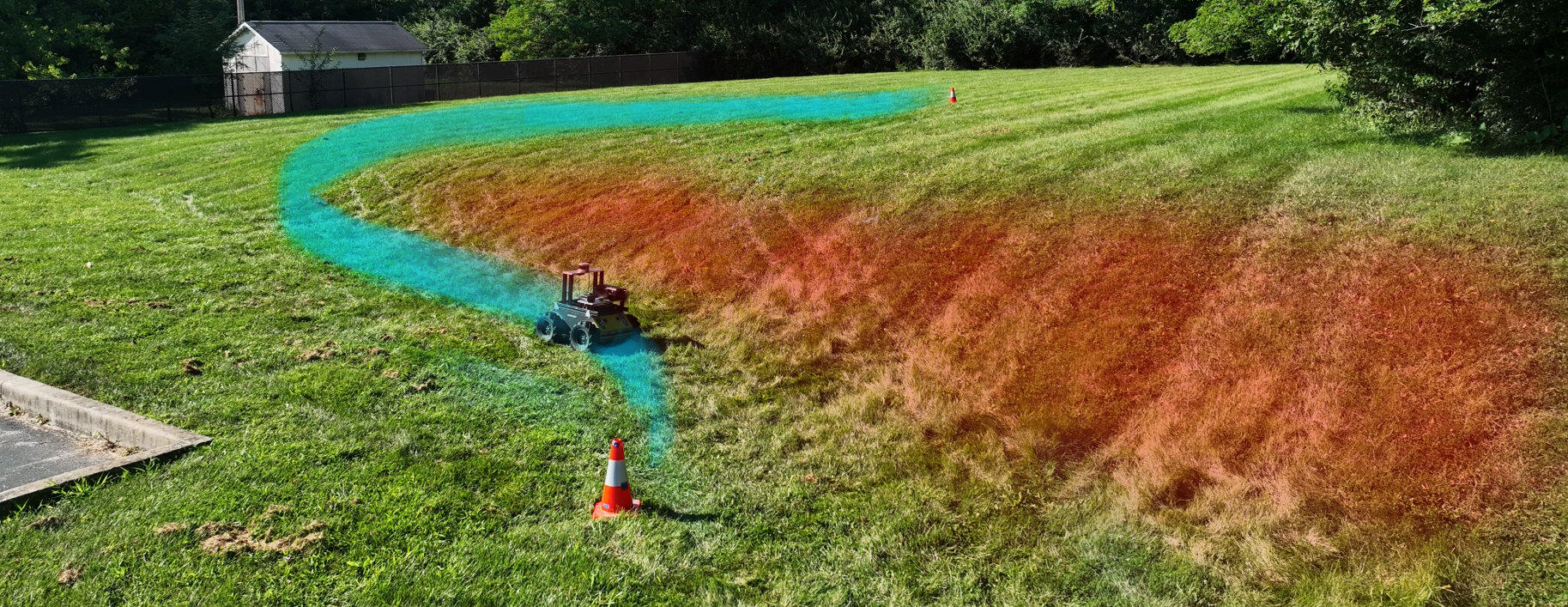}} 
\\ 
\subfloat[Hilly Terrain Environment (HT)\label{hilly_terrain}]{%
  \includegraphics[width=0.45\textwidth,height=0.8in]{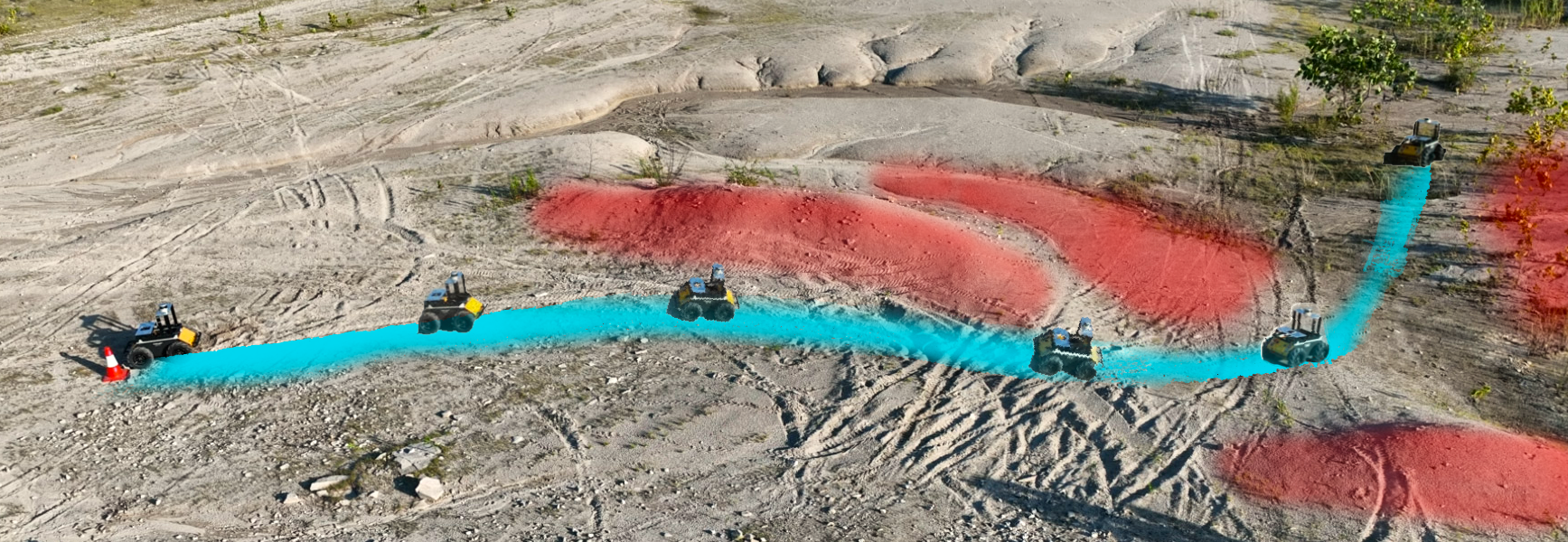}} 
\caption{\small Real-world demonstration environments. The paths taken by the robot are highlighted in blue while the hills are highlighted in red. Video clips demonstrating the robot's performance can be found in the supplementary materials.}
\vspace{-8pt}
\label{real_word_exp}
\end{figure}

\section{Real-world Demonstrations}\label{Real-world Demonstrations}
We experimentally demonstrate that our proposed approach can navigate a real robot through uneven terrain safely. 
\subsection{Setup}
For the experimental validation, we used the simulation setup previously outlined in Section~\ref{Simulation Setup}, except
for that the Velodyne is used for robot localization using a \textit{GPS}, and the robot's maximum linear velocity was limited to  $v_{max}$ to $0.8 m/s$.
(We opted not to implement the baseline in the real-world context due to its requirement for an exhaustive scan and meticulous map building of the demonstration site, which is a strong assumption that is not practical in many applications and also a tedious task beyond our present focus.)

Two environments are considered for the real-world demonstration, namely, the \textit{Gentle Slope Bypass} (GSB) environment and the \textit{Hilly Terrain} (HT) environment, see Fig.~\ref{real_word_exp}. The \textit{GSB} environment features a small grassy hill; for an optimal path to the goal, the robot should circumvent the hill until it is safe to climb, then the robot should navigate towards the goal. The \textit{HT} environment presents more challenging terrains, where the robot needs to avoid a hilly area to reach its destination. 

\subsection{Results}
 The performance of the proposed approach during the real-world demonstration in both environments, \textbf{GSB} and \textbf{HT}.
The performance statistics show that our approach successfully navigates the robot in the tested environments. In \textbf{GSB}, the maximum roll angle $\phi_{max}=0.2 rad$ exceeds the maximum pitch angle $\psi_{max} = 0.12 rad$. This is attributed to the robot's trajectory; it traversed the lowest part of the slope diagonally until reaching a section where it was feasible to curve towards the goal, as seen in \ref{gentle_slope_bypass}. The average change in elevation $\dot{z}_{r_{avg}}$ is $2 cm/s$ and the total traveled distance is around $50m$. 
In \textbf{HT}, the terrain's incline aligns with the robot's direction, hence the pitch angle $\psi_{max} = 0.2rad$ predominantly exceeds the roll angle $\phi_{max}=0.15 rad$. This scenario also records an average elevation change of $\dot{z}_{r_{avg}}$ of $1.1cm/s$ and a traveled distance of $29m$. The observed low vibration rates signify the stability of our method. The high averages curvature changes, denoted as ${C}_{chg}$, can be attributed to our mapless approach where subgoal placements are determined by the real-time readings from the LiDAR, without the incorporation of a global planner.


\newcolumntype{P}[1]{>{\centering\arraybackslash}p{#1}}

\section{Conclusion and Future Work}\label{sec:conclusion}
This paper proposes a new mapless navigation approach for robots operating in uneven terrain. The approach utilizes an efficient variant of Sparse Gaussian Process to learn the steepness and flatness of the robot's surrounding environment. The simulation and real-world experiments' results have shown that the proposed algorithm is able to navigate the robot in uneven terrain without the need for a global map. 
In forthcoming research endeavors, we intend to refine the methodology by incorporating obstacle consideration within navigable paths, enhancing the cost function's responsiveness to negative slopes such ditches, and integrating an optimization-based controller for improved navigation efficiency.
\bibliographystyle{IEEEtran}
\bibliography{references}            


\end{document}